%% file: 0_main.tex
\documentclass[sigconf, 9pt,twocolumn]{acmart}
\makeatletter
\def\@ACM@checkaffil{
    \if@ACM@instpresent\else
    \ClassWarningNoLine{\@classname}{No institution present for an affiliation}%
    \fi
    \if@ACM@citypresent\else
    \ClassWarningNoLine{\@classname}{No city present for an affiliation}%
    \fi
    \if@ACM@countrypresent\else
        \ClassWarningNoLine{\@classname}{No country present for an affiliation}%
    \fi
}
\makeatother

\usepackage{xspace}
\usepackage{enumitem}
\usepackage{graphicx}
\usepackage{subcaption}
\usepackage{multirow}
\usepackage{balance}
\usepackage{enumitem}
\usepackage{siunitx}

\input{preambles}
\begin{document}

\title{Exploring the Capabilities of LLMs for IMU-based Fine-grained Human Activity Understanding}

\author{Lilin Xu$^\ast$, Kaiyuan Hou$^\ast$, Xiaofan Jiang}
\affiliation{\institution{Columbia University}}
\email{lx2331@columbia.edu, kh3119@columbia.edu, jiang@ee.columbia.edu}
\thanks{$\ast$ Equal contribution.}

\def \authors{Lilin Xu, Kaiyuan Hou, Xiaofan Jiang}
\renewcommand{\shortauthors}{Lilin Xu, Kaiyuan Hou, Xiaofan Jiang}

\input{tex_fmsys/00_abstract.tex}

\copyrightyear{2025} 
\acmYear{2025} 
\setcopyright{acmlicensed}\acmConference[FMSys '25]{The 2nd International
Workshop on Foundation Models for Cyber-Physical Systems \& Internet of
Things }{May 6--9, 2025}{Irvine, CA, USA}
\acmBooktitle{The 2nd International Workshop on Foundation Models for
Cyber-Physical Systems \& Internet of Things (FMSys '25), May 6--9, 2025,
Irvine, CA, USA}
\acmDOI{10.1145/3722565.3727195}
\acmISBN{/2025/05}



\maketitle

\input{tex_fmsys/01_introduction}

\input{tex_fmsys/03_prelim}
\input{tex_fmsys/04_finetune}

\input{tex_fmsys/05_mapping}
\input{tex_fmsys/07_discussion}

\input{tex_fmsys/06_conclusion}

\input{tex_fmsys/ack}
\balance

\bibliographystyle{ACM-Reference-Format}
\bibliography{reference}
\newpage

\end{document}

%% file: tex_fmsys/00_abstract.tex
\begin{abstract}
Human activity recognition (HAR) using inertial measurement units (IMUs) increasingly leverages large language models (LLMs), yet existing approaches focus on coarse activities like walking or running. Our preliminary study indicates that pretrained LLMs fail catastrophically on fine-grained HAR tasks such as air-written letter recognition, achieving only near-random guessing accuracy. In this work, we first bridge this gap for flat-surface writing scenarios: by fine-tuning LLMs with a self-collected dataset and few-shot learning, we achieved up to a $129\times$ improvement on 2D data. To extend this to 3D scenarios, we designed an encoder-based pipeline that maps 3D data into 2D equivalents, preserving the spatiotemporal information for robust letter prediction. Our end-to-end pipeline achieves 78\% accuracy on word recognition with up to 5 letters in mid-air writing scenarios, establishing LLMs as viable tools for fine-grained HAR.
\end{abstract}

%% file: tex_fmsys/01_introduction.tex
\section{Introduction}\label{sec: intro}

\begin{figure}[t]
    \centering
    \includegraphics[width=\linewidth]{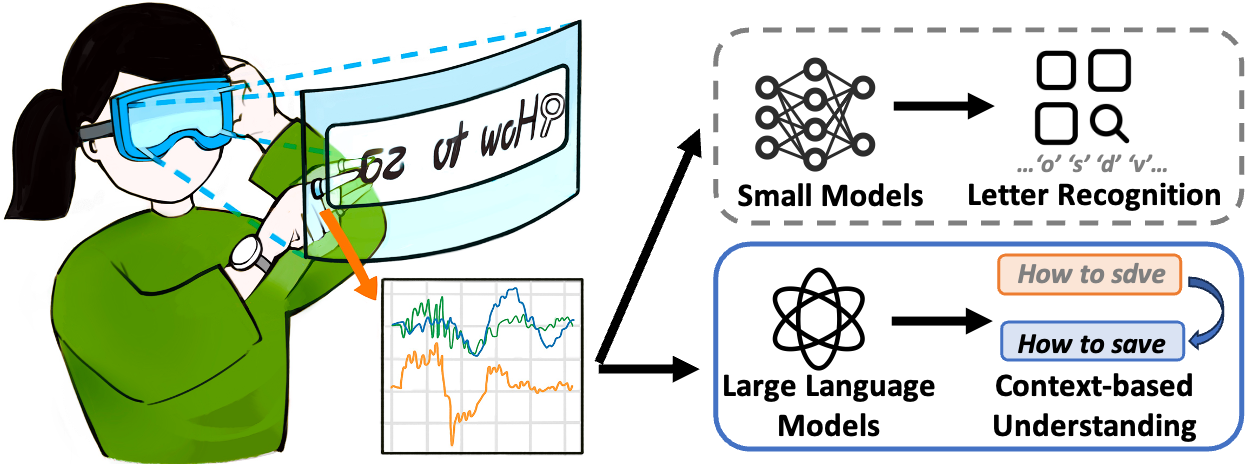}
    \caption{Potential application scenarios of LLM-based fine-grained human activity understanding. Compared with small models, LLMs can enhance various context-based applications with their strong contextual understanding and generalization capabilities.}
    \label{fig:application}
\end{figure}

Large Language Models (LLMs)~\cite{ouyang2022training, achiam2023gpt, dubey2024llama} have gained significant attention due to their exceptional generalization and minimal adaptation required to new domains across diverse downstream tasks, including healthcare~\cite{yang2024drhouse,he2024myo,liu2023large}, monitoring~\cite{yu2025sensorchat,liu2024tasking,chen2024sensor2text}, and assistive systems~\cite{guo2024vivar, yang2025socialmind,liu2024tasking}. This adaptability makes these models particularly promising for sensor-based applications~\cite{xu2024penetrative,reichman2025sensorqa}, where traditional approaches often require domain-specific feature engineering~\cite{xu2023mesen,xu2024gestureprint}. Building on this trend, researchers have begun exploring how to leverage LLMs for Inertial Measurement Unit (IMU)-based Human Activity Recognition (HAR). This requires LLMs to interpret complex motion patterns from multiple IMU channels, which presents unique challenges in leveraging the contextual understanding capabilities of large models.

Researchers have primarily explored two approaches for applying LLMs to IMU-based HAR. The first, \textbf{Knowledge-Driven HAR with Pre-Trained LLMs}, employs semantic grounding to map spatiotemporal IMU data to textual descriptions of activities~\cite{ji2024hargpt, ouyang2024llmsense}. This method relies on the world knowledge already learned in LLMs, enabling recognition of various activity classes without specific training. With proper prompt engineering, LLMs could provide satisfactory answers on several open-source datasets. However, it struggles with ambiguities when mapping raw IMU signals to precise textual labels. For instance, distinguishing subtle differences such as \textit{stair ascent vs. descent} remains challenging due to the lack of explicit IMU-to-language alignment.

The second approach focuses on bridging this semantic gap, which usually achieves better performance and is more resilient to sensor noises~\cite{imran2024llasa, yan2024language, li2024sensorllm}. This approach requires additional training and can be broadly categorized into two types: (1) \textbf{Modality-Alignment}, which is a projection-based fusion, where lightweight encoders map IMU data to the LLM's text embedding space. (2) \textbf{Fine-Tuning} that treats time-series sensor data as pseudo-text inputs for direct processing by LLMs. 
While challenges such as computational overhead and dependency on paired text-IMU datasets remain active areas of optimization, this approach shows promise for real-world deployment. Nevertheless, current research predominantly targets coarse-grained activities (e.g., \textit{walking, sit-stand, jogging}), leaving fine-grained activity recognition largely unexplored.

Fine-grained HAR is one of the key problems in mobile sensing applications, particularly for domains requiring detailed gesture understanding such as AR/VR applications, which currently mainly rely on camera-based tracking solutions, like HTC Vive~\cite{htcvive} and Meta Quest~\cite{metaquest}. However, this solution is limited by light conditions and often fails to work during nights or in environments with poor visibility. IMU sensors offer a promising alternative to these vision-based solutions by providing motion data that works reliably in challenging conditions like low lighting or occlusion. While existing works have successfully used machine learning approaches to gesture recognition, these methods often require extensive labeled datasets specific to each application domain. Besides, they suffer from distribution shifts~\cite{ji2024hargpt} caused by varying device types, sensor placements, and subject differences. As a result, traditional HAR models have severe performance degradation when applied to unseen data from new environments or users, even when recognizing the same activities.
This is where LLMs show particular promise.

In this work, we are focusing on this challenging fine-grained HAR problem: recognizing letters drawn in mid-air using only IMU data. This 26-class classification task represents a substantial increase in complexity over typical coarse-grained activities found in existing literature. Unlike walking or sitting, which have distinct motion signatures, letters like `O' and `Q' or `B' and `R' involve subtle trajectory differences. The complexity is further compounded by the free-form nature of mid-air writing, where users lack the tactile feedback and physical constraints present in traditional writing surfaces, leading to higher variability. Besides, the temporal dynamics of letter drawing (varying speeds, pauses, and stroke orders across users) make it even challenging.  Our contributions include:
\begin{itemize}[leftmargin=*]
    \item \textbf{Zero-shot vs. few-shot analysis:} LLMs perform below random guessing in zero-shot settings for fine-grained HAR tasks. Few-shot learning only benefits models already capable of analyzing time-series data, while fine-tuning enables initially poor-performing models to provide meaningful analysis.
    
    \item \textbf{A dimensional mapping framework:} LLMs can recognize flat-surface 2D gestures but consistently fail with mid-air 3D data. Our mapping approach successfully translates 3D IMU data to their 2D equivalents, addressing a limitation in LLM-based gesture recognition.
    
    \item \textbf{An End-to-end pipeline for word recognition:} The proposed pipeline, from 3D IMU input to word prediction, achieves 78\% accuracy for up to 5 letters, demonstrating practical viability for fine-grained HAR applications in realistic settings.
\end{itemize}

%% file: tex_fmsys/03_prelim.tex
\section{Preliminary}\label{sec: prelim}
In this section, we first collect a dataset of letters using a device equipped with an IMU sensor, capturing data both on a flat surface and in mid-air.
Then we evaluate the recognition performance of pretrained LLMs with both zero-shot and few-shot settings, indicating the potential and challenges of LLMs in fine-grained HAR.

\subsection{Data Collection}\label{sec:data}

\begin{figure}[t]
    \centering
    \captionsetup{skip=3pt}
    \begin{subfigure}[b]{0.23\textwidth}
        \centering
        \includegraphics[width=\linewidth]{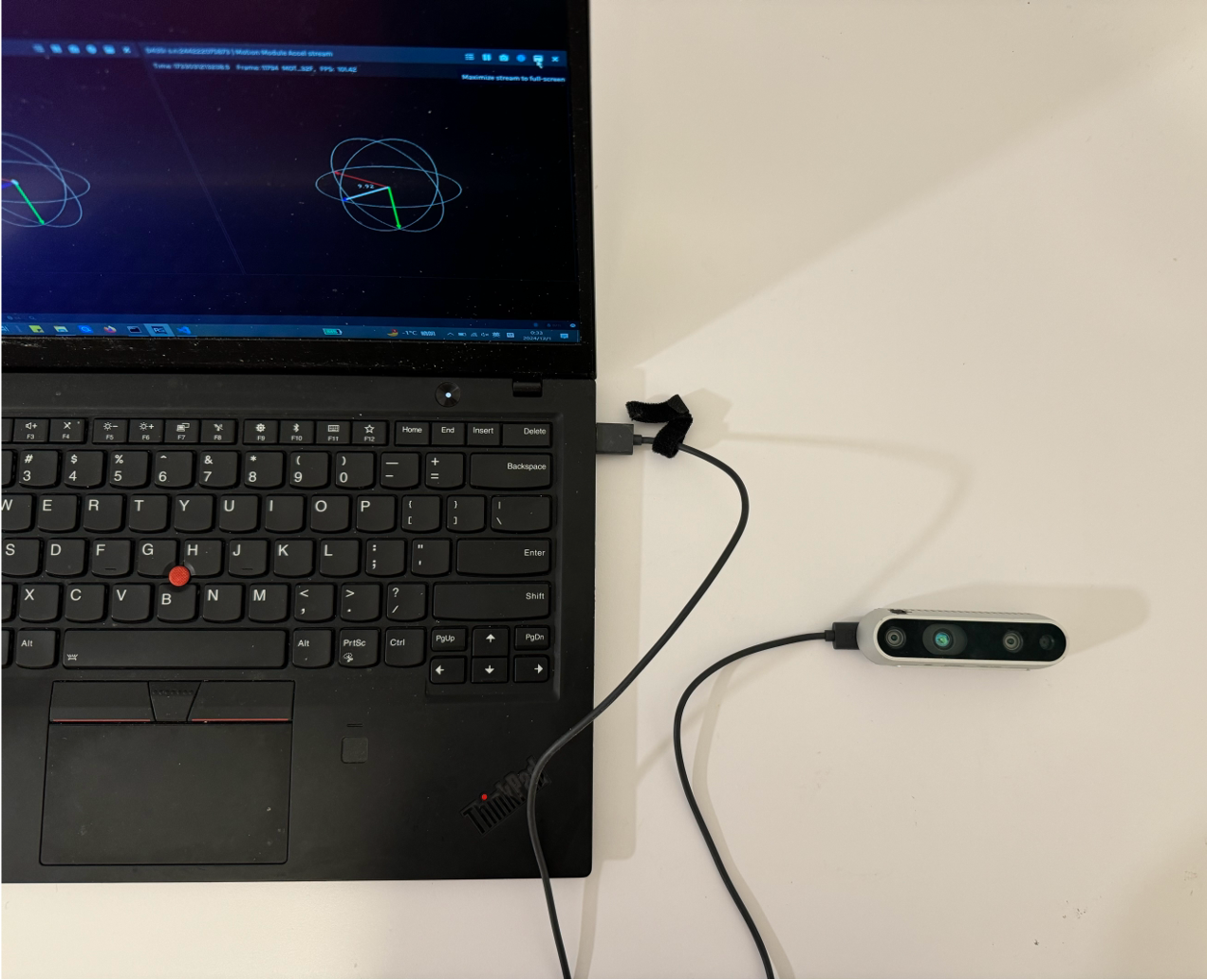}
        \caption{The collection setup.}
        \label{fig:collection_env}
    \end{subfigure}
    \hfill
    \begin{subfigure}[b]{0.23\textwidth}
        \centering
        \includegraphics[width=\linewidth]{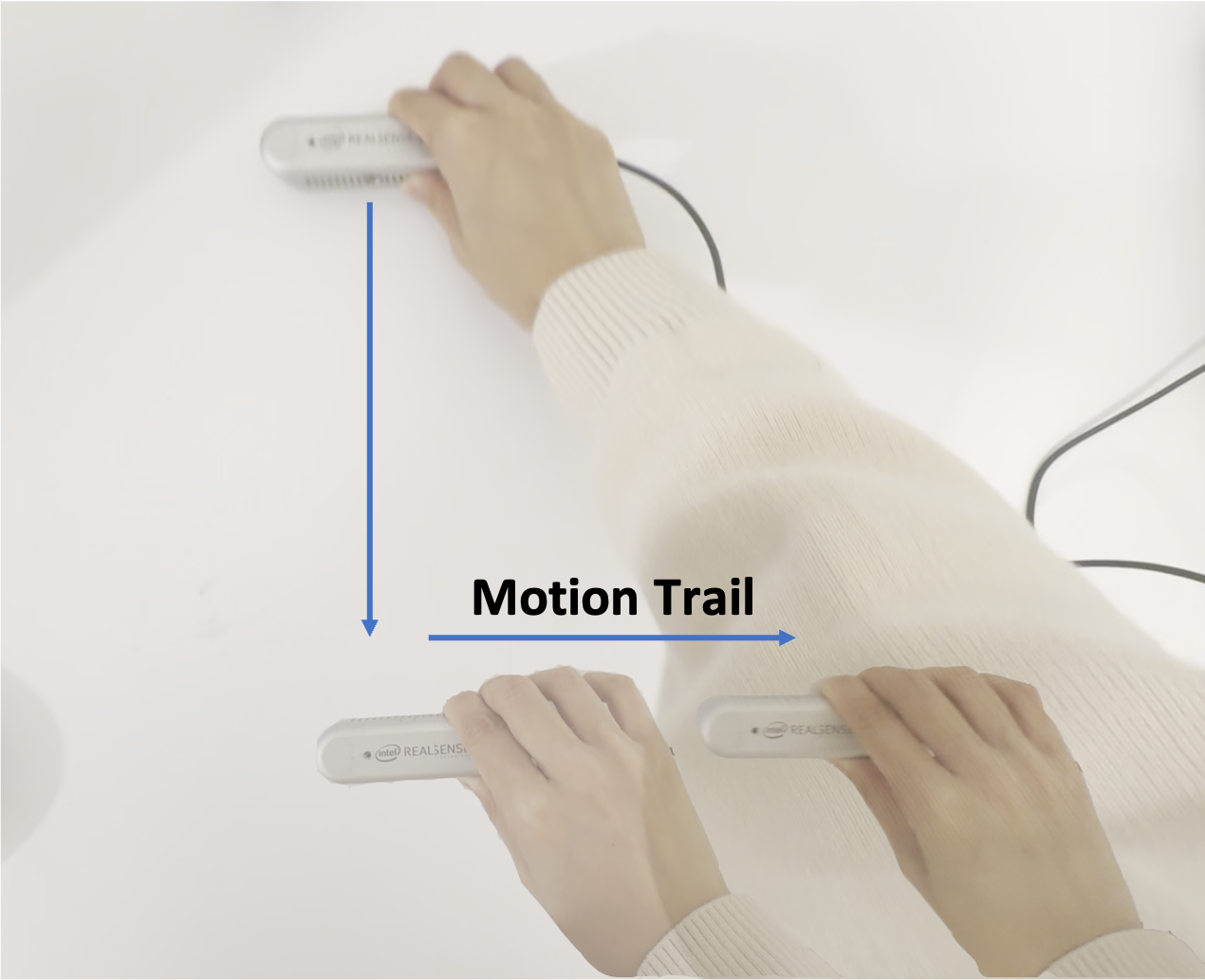}
        \caption{The motion trail of `L'.}
        \label{fig:2d_motion}
    \end{subfigure}
    \caption{
    The data collection setup and process.}
    \label{fig:data_collection}
\end{figure}

We utilized the Intel RealSense Depth Camera D435i~\cite{realsense}, equipped with a Bosch BMI055 inertial sensor, to collect data. This sensor provides measurements of linear acceleration, enabling the capture of detailed motion information.
We collected data from two distinct scenarios: writing on a flat surface and writing in mid-air, referred to as 2D and 3D cases, respectively. 
The collection setup and process are illustrated in Figure \ref{fig:data_collection}.
As shown in Figure \ref{fig:collection_env}, the collection setup includes the sensor and a laptop.
Figure \ref{fig:2d_motion} shows the trajectory of the device during the collection of the letter `L'.

\begin{figure}[t]
    \centering
        \includegraphics[width=0.9\linewidth]{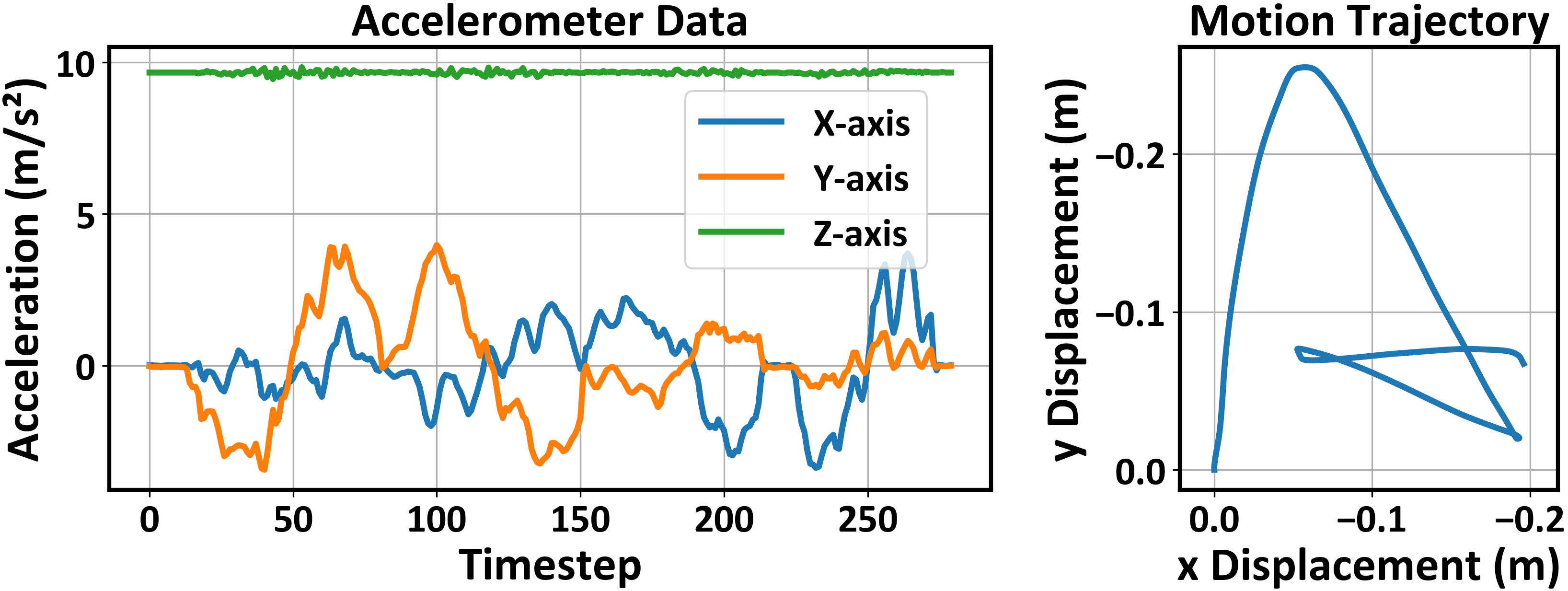}
    \caption{Visualization of collected IMU data (2D).}
    \label{fig:letter_vis}
\end{figure}

In both 2D and 3D cases, each participant drew all 26 letters 10 times repeatedly, resulting in 520 samples per person (26 letters $\times$ 2 cases $\times$ 10 repetitions). With data collected from two participants, our complete dataset contains a total of 1,040 samples.
Data from one participant is used as the training set, while data from the other participant serves as the test set.
An example of the collected data for the letter `A' in 2D is shown in Figure~\ref{fig:letter_vis}.
The left portion in the figure shows the three-axis accelerometer data from the sensor, while the right portion represents the reconstructed letter based on the time-series data.

\subsection{Microbenchmark}
With the collected dataset, we do a preliminary study to evaluate the recognition performance of pretrained LLMs. 

\noindent\textbf{Baselines.} We employ classical machine learning models (Random Forest~\cite{biau2016random}, SVM~\cite{hearst1998support}) and deep learning models (DCNN~\cite{yang2015deep}, LIMU-LSTM~\cite{xu2021limu}) as baseline models.
To ensure a fair comparison with LLMs, we applied the same x-way 1-shot settings to the baseline models as we will use for LLMs. Specifically, we randomly sampled one instance of each letter (26 in total) from both the 2D and 3D cases from the training set, resulting in 52 training samples. The models were then evaluated on the test set. We acknowledge that this is a constrained training scenario compared to conventional fully-supervised approaches, but this limitation enables direct comparison with few-shot LLM performance. The results are the average of five runs.

\begin{figure}[t]
    \centering  
    \includegraphics[width=\linewidth]{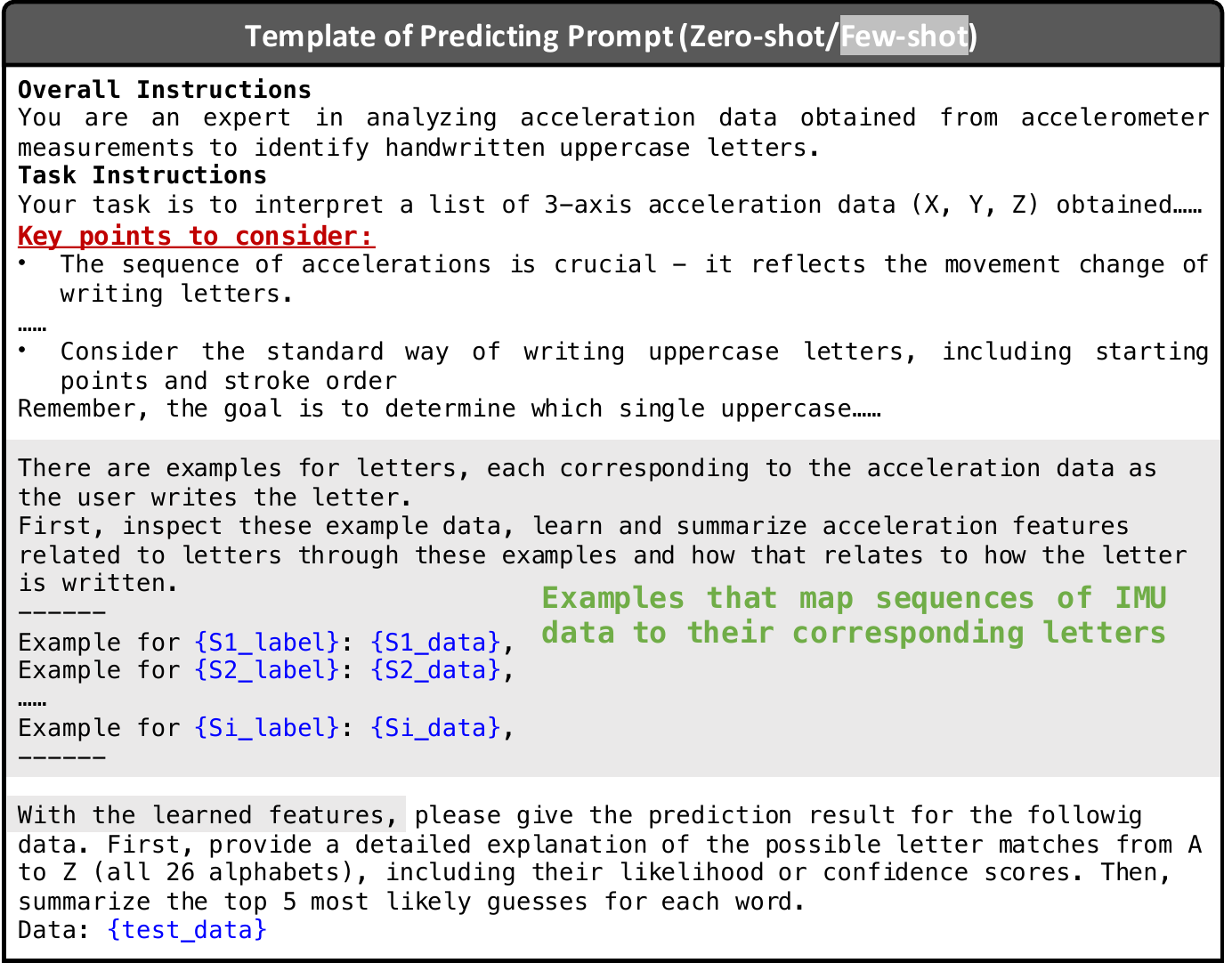}
    \caption{Prompt templates. The additional context specific to the few-shot template is highlighted in gray.}
    \label{fig:preexp_prompt}
\end{figure}

\noindent\textbf{LLMs.} We employ three LLMs (LLaMA-3-8B~\cite{llama3_8b}, GPT-4o~\cite{gpt4o}, and DeepSeek-R1~\cite{deepseek}) in both zero-shot and few-shot settings as shown in Figure~\ref{fig:preexp_prompt}. In these prompts, we incorporate Chain-of-Thought (CoT)~\cite{wei2022chain} for improving LLMs' reasoning to process complex tasks and follow instructions effectively.
For the few-shot setting, we provide several examples of IMU data paired with their corresponding letters. Unlike the training of baseline models where we included one sample per class, the context size limitations of LLMs prevent us from including examples for all 26 letters. Instead, we randomly select a subset of examples from the training samples used in baseline model training. We evaluate the LLMs separately on 2D and 3D data from the test set to assess their performance in each scenario.

\noindent\textbf{Results.}
Table \ref{tab:comparison} shows the recognition performance of baseline models and LLMs. Our analysis reveals several key insights: In \textbf{{zero-shot settings}}, all evaluated LLMs performed poorly, with accuracies falling below random guessing (3.85\% for a 26-class problem). This suggests that despite their sophisticated reasoning capabilities in other domains, LLMs without examples lack the intrinsic ability to interpret raw IMU signals and map them to corresponding letter shapes. When provided with examples in \textbf{few-shot settings}, we observed notable performance differentiation:
\begin{itemize}[leftmargin=*]
    \item On 2D letter drawing data, GPT-4o achieved approximately 26.5\% accuracy, comparable to other baselines.
    \item DeepSeek-R1, a state-of-the-art reasoning model, achieved 59.6\% accuracy on 2D data, substantially outperforming both baseline models and other LLMs.
    \item Smaller models, specifically LLaMA-3-8B, failed to show improvement even with examples provided, primarily due to their inability to properly interpret this time-series classification task, as illustrated in Figure~\ref{fig:pretrainedllama_output}.
\end{itemize}
Notably, \textbf{all LLMs performed poorly on 3D letter recognition} regardless of the experimental setting, indicating that mid-air gestures present additional complexities that current LLMs struggle to interpret. Pretrained LLMs cannot directly handle fine-grained HAR tasks due to the fundamental differences between IMU data representation and the text data they were trained on. Thus, in the following section, we will inject this expert knowledge by fine-tuning to see if this can help improve their performance.

\begin{table}[t!]
    \centering
    \renewcommand{\arraystretch}{1.0}
    \begin{tabular}{c c c c}
        \toprule
        \multirow{2}{*}{\textbf{Method}} & \multirow{2}{*}{\textbf{Setting}} & \multicolumn{2}{c}{\textbf{Evaluation Metric}} \\
        \cmidrule(lr){3-4}
        & & \textbf{Accuracy} & \textbf{F1-Score} \\
        \midrule
        RF &  Supervised & 0.259 & 0.252 \\
        SVM & Supervised & 0.304 & 0.288 \\
        DCNN & Supervised & 0.254 & 0.245 \\
        LIMU-LSTM & Supervised & 0.281 & 0.243 \\
        \midrule
        \multirow{2}{*}{LLaMA-3-8B} & Zero-shot & 0 | 0 & 0 | 0  \\
        & Few-shot & 0.004 | 0 & 0.003 | 0  \\
        \midrule
        \multirow{2}{*}{GPT-4o} & Zero-shot & 0.042 | 0.019 & 0.046 | 0.017  \\
        & Few-shot & 0.265 | 0.046 & 0.263 | 0.042  \\
        \midrule
        \multirow{2}{*}{DeepSeek-R1} & Zero-shot & 0.031 | 0.031  & 0.029 | 0.028 \\
        & Few-shot & 0.596 | 0.054 & 0.590 | 0.051 \\
        \bottomrule
    \end{tabular}
    \caption{Recognition performance of baselines and LLMs in zero/few-shot settings. 
    The results of LLMs are split for 2D and 3D cases with vertical bars [`(2D)|(3D)'].
    }
    \label{tab:comparison}
\end{table}

%% file: tex_fmsys/04_finetune.tex
\section{LLM Adaptation for IMU-based Letter Recognition}\label{sec: adaption}
For the task of classifying letters from IMU data, we fine-tune two instruction-following language models: GPT-4o and LLaMA-3-8B. This requires constructing appropriate instruction-response pairs tailored for this classification task. Since our collected IMU data is not naturally formatted as instructions or responses, we first convert the dataset collected in \S\ref{sec:data} into instruction-answer pairs~(\S\ref{sec:instruction_answer}) suitable for LLM fine-tuning (\S\ref{sec: finetune}). Then, we evaluate fine-tuned LLMs on single letter prediction (\S\ref{sec:exp1}).

\subsection{Instruction-Answer Pair Generation}
\label{sec:instruction_answer}
As demonstrated in \S~\ref{sec: prelim}, simply inputting raw IMU data and requesting classification yields poor results. To address this limitation and automate the creation of training samples, we implemented a multi-step approach to generate an instruction-response dataset that can be used to fine-tune LLMs for letter recognition.

\subsubsection{Generating Self-Consistent Explanations}
We paired each IMU sample with its ground truth letter from the training set, and then prompted an LLM to generate explanations for why the IMU data corresponds to that specific letter. This step enables the LLM to construct plausible reasoning based on the provided ground truth. 

\subsubsection{Separating Reasoning from Conclusions}
We employed another LLM to restructure these initial responses into clear explanations that separate the reasoning process from the conclusion. The separated reasoning will show the natural cognitive process of inferring letters from motion patterns and serves as the response component of our training pairs.

\subsubsection{Diversifying Instruction Prompts}
For the instruction part, we appended the IMU data with a predefined prompt template and used a third model to generate variations of this prompt to make the training data more diverse. We repeated this process three times for each sample in our dataset, resulting in 1,560 instruction-response pairs for model fine-tuning.

\subsection{Model Fine-tuning}\label{sec: finetune}
We fine-tuned two models, LLaMA-3-8B and GPT-4o, with LoRA~\cite{hu2021lora}. Each model was trained on the instruction-response dataset for 5 epochs with a learning rate of 1e-5. Through this fine-tuning process, we would like to investigate the following three questions:
\begin{enumerate}[leftmargin=*]
    \item Can fine-tuning enable underperforming models like LLaMA-3-8B to effectively interpret IMU data?
    \item Does fine-tuning further enhance the performance of state-of-the-art models, or is there an inherent ceiling to the accuracy achievable by LLMs on this task?
    \item Can fine-tuning improve model performance on the more challenging 3D letter recognition task?
\end{enumerate}
\subsection{Exp: Single Letter Prediction}\label{sec:exp1}
We used the test set, which is collected by the other person as mentioned in Section~\ref{sec:data}, to evaluate fine-tuned LLMs. The results presented in Table~\ref{tab:model_comparison} reveal several notable insights as follows.

\begin{table}[t!]
\centering
\small
\resizebox{1.\linewidth}{!}{
\begin{tabular}{ccc|cc}
\hline
\multirow{2}{*}{\textbf{Model}} & \multicolumn{2}{c|}{\textbf{Zero-shot}} & \multicolumn{2}{c}{\textbf{Few-shot}} \\
 & \textbf{Pretrained} & \textbf{Fine-tuned} & \textbf{Pretrained} & \textbf{Fine-tuned} \\
 \hline
LLaMA-3-8B  & 0 | 0 & 3.46 | 2.69  & 0.38 | 0    & 49.23 | 5.77 \\
\hline
GPT-4o  & 4.23 | 1.92  & 6.15 | 3.08  & 26.54 | 4.62  & 58.08 | 6.54 \\
\hline
\end{tabular}}
\caption{
Recognition accuracy (\%) of GPT-4o and LLaMA-3-8B, with 2D and 3D cases split by vertical bars.
}
\label{tab:model_comparison}
\end{table}

\begin{figure}[t]
    \centering
    \captionsetup{skip=3pt}
    \begin{subfigure}[b]{0.48\textwidth}
        \centering
        \captionsetup{skip=3pt}
        \includegraphics[width=\linewidth]{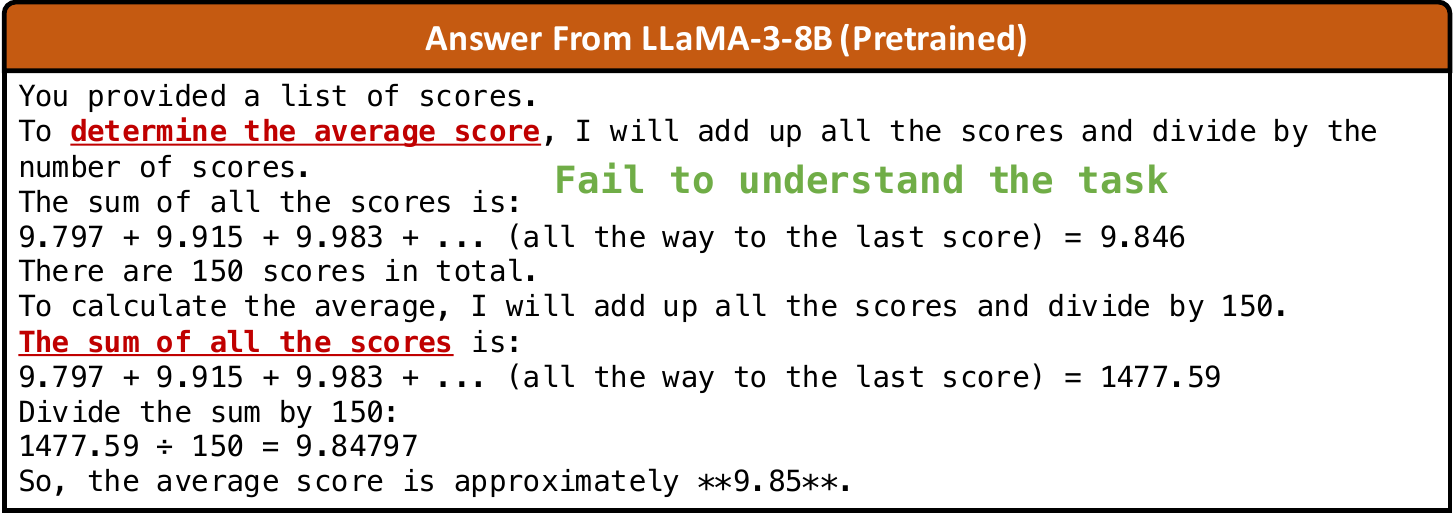}
        \caption{An example of pretrained LLaMA-3-8B’s answer, indicating the pretrained one fails to understand the task.}
        \label{fig:pretrainedllama_output}
    \end{subfigure}
    \hfill
    \begin{subfigure}[b]{0.48\textwidth}
        \centering
        \captionsetup{skip=3pt}
        \includegraphics[width=\linewidth]{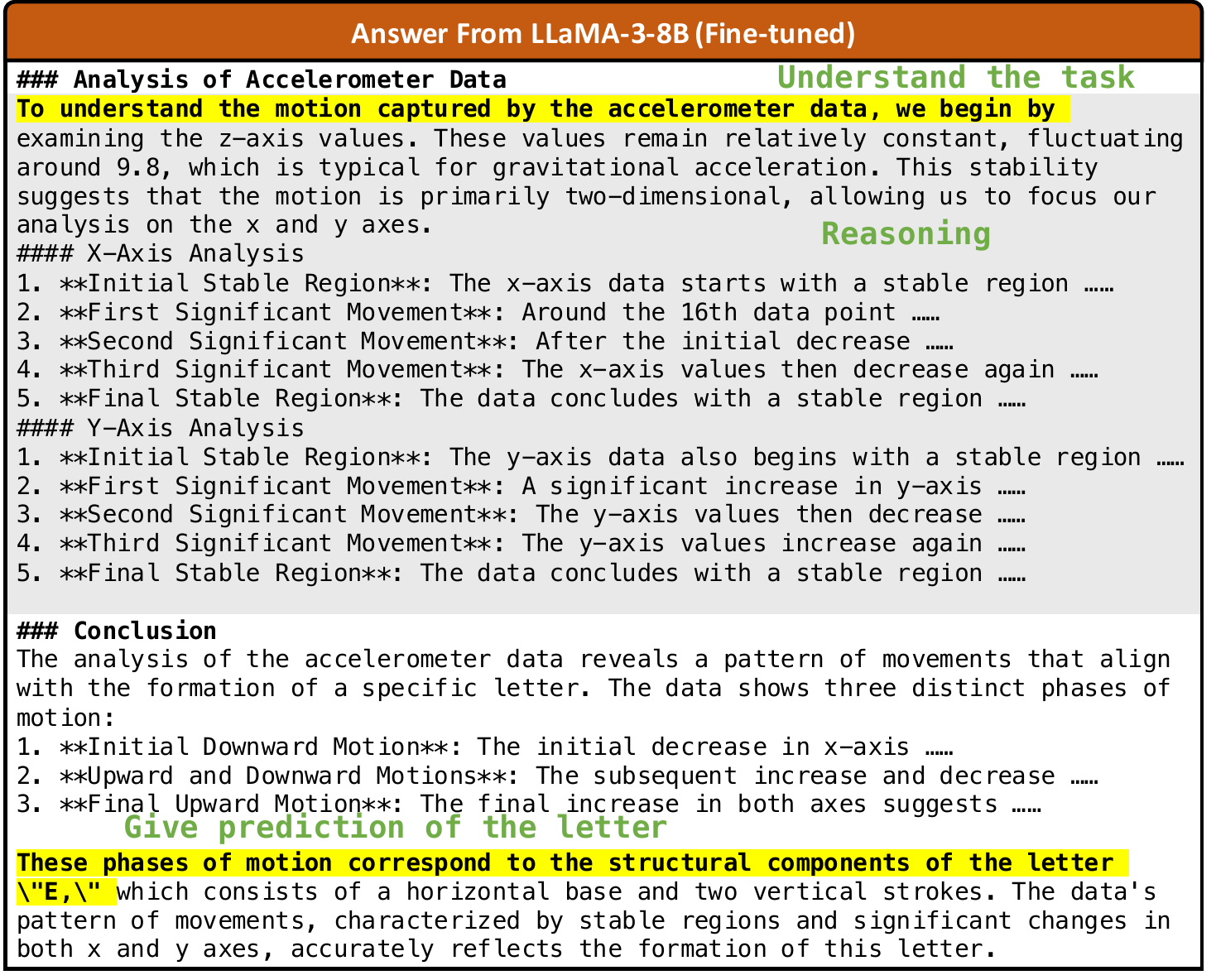}
        \caption{An example of fine-tuned LLaMA-3-8B’s answer, indicating the fine-tuned one can perform letter recognition.}
        \label{fig:finetunedllama_output}
    \end{subfigure}
    \caption{Examples of LLaMA-3-8B models' answers.}
    \label{fig:llama_output}
\end{figure}

\textit{Fine-tuning improves performance across both models.} 
Fine-tuning consistently enhances the performance of both models, though with varying degrees of improvement. GPT-4o shows notable improvements in accuracy after fine-tuning, particularly in the few-shot setting and achieved similar accuracy with DeepSeek-R1. Similarly, fine-tuning significantly benefits LLaMA-3-8B, a model that initially failed at this task, as the response shown in Figure~\ref{fig:finetunedllama_output}. With few examples provided, LLaMA-3-8B's recognition accuracy for 2D data in the few-shot setting dramatically improved from 0.38\% to 49.23\%, yielding a remarkable 129× increase. This indicates that fine-tuning can effectively teach models to interpret IMU data patterns that they were previously unable to recognize.

\begin{figure}[t]
    \centering
    \includegraphics[width=0.65\linewidth]{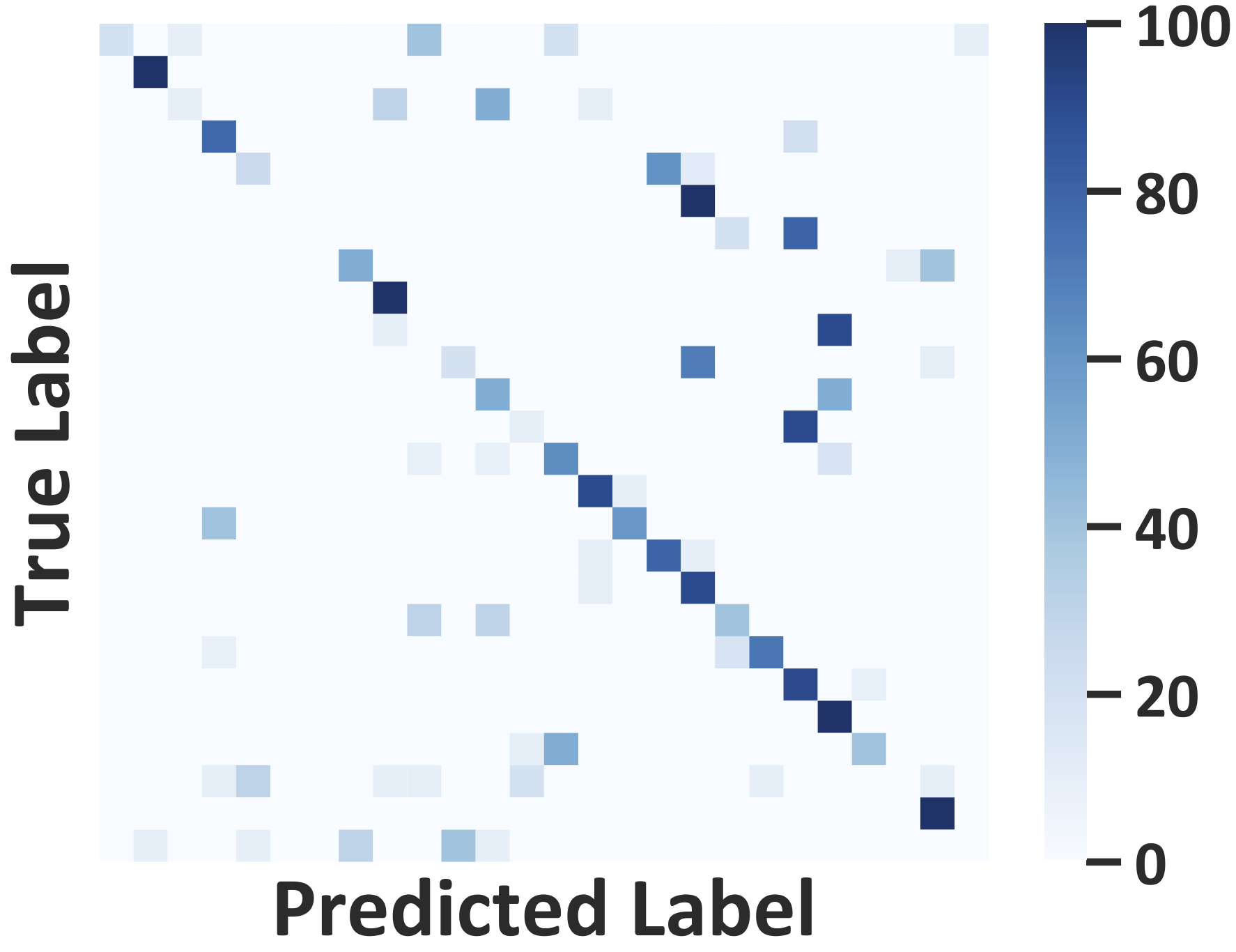}
    \caption{
    Confusion Matrix of fine-tuned LLaMA-3-8B in the few-shot setting on the test set.}
    
\label{fig:llama_confusion_test}
\end{figure}

\textit{Few-shot learning substantially improves accuracy.} 
Providing examples that explain how IMU data corresponds to specific letters results in significant accuracy improvements for most models, with an up to $14\times$ improvement on LLaMA-3-8B. The confusion matrix for the fine-tuned LLaMA-3-8B model on the test set in a few-shot learning setting is shown in Figure~\ref{fig:llama_confusion_test}. Both the x-axis and y-axis ticks represent the 26 letters of the alphabet. Notably, models that initially cannot understand the task, such as the pretrained LLaMA-3-8B, cannot benefit from simply providing examples, producing only one correct result out of 260 tests.

\textit{Performance on 3D data remains poor.} 
Despite improvements in 2D cases, both models still yield very limited accuracy in 3D cases, which represent more realistic scenarios. This persistent challenge suggests fundamental limitations in current LLM approaches to mid-air gesture recognition.

%% file: tex_fmsys/05_mapping.tex
\begin{figure}[t]
    \centering
        \includegraphics[width=\linewidth]{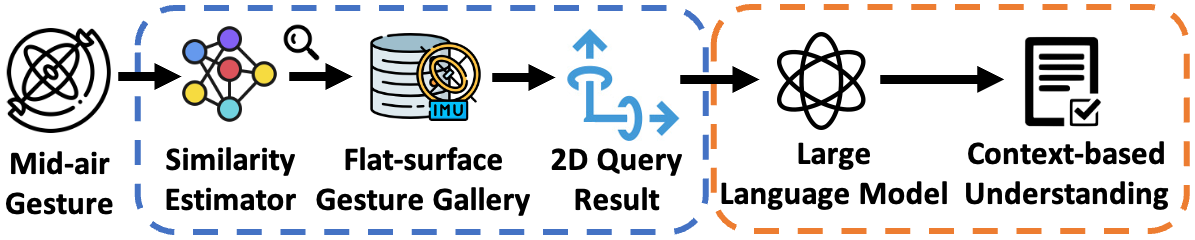}
    \caption{Overview of proposed end-to-end mid-air gesture understanding pipeline based on LLMs.}
    \label{fig:overview}
\end{figure}

\section{Spatial Motion Mapping}\label{sec: mapping}
To address the challenge of recognizing 3D IMU data, we developed a pipeline that maps 3D IMU data to 2D representations that can be interpreted by LLMs. 
This method serves as a preprocessing step, converting the 3D IMU data into 2D that retains the essential features of the letters while being more accessible to LLMs' interpretive capabilities. Once mapped to 2D, the data can be processed by LLMs to perform fine-grained HAR understanding. Figure \ref{fig:overview} shows the overview of this pipeline.

\subsection{Dimensional Reduction}
\begin{figure}[t]
    \centering
        \includegraphics[width=\linewidth]{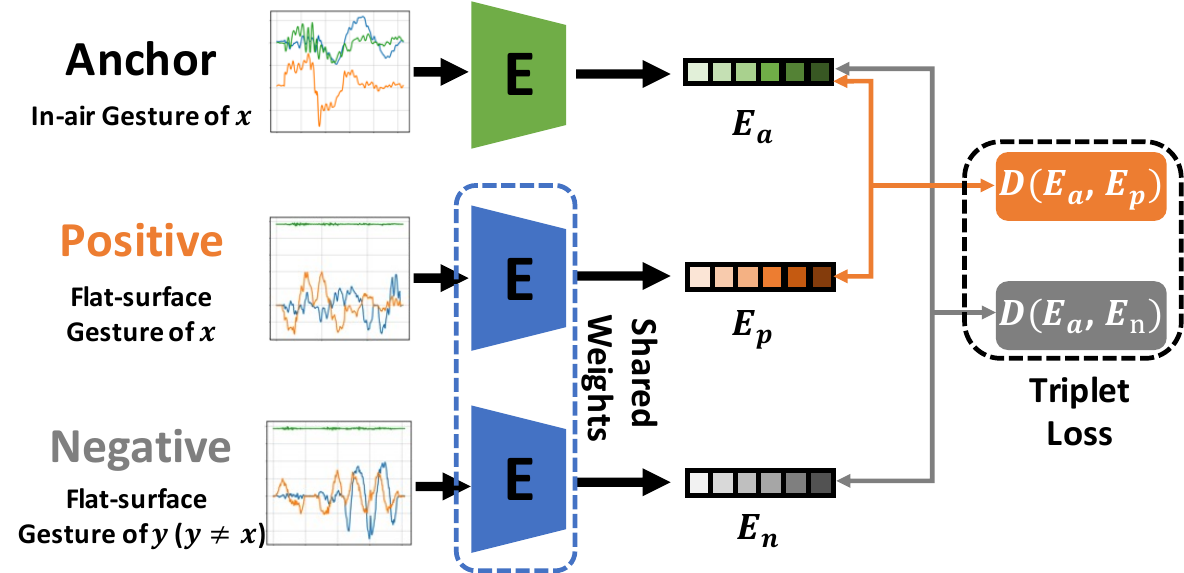}

    \caption{The framework of mapping mid-air gestures to flat-surface gestures through deep metric learning.}
    \label{fig:map_cl}
\end{figure}

We employed deep metric learning~\cite{ge2018deep} to map 3D IMU data to 2D data using a triplet architecture as illustrated in Figure~\ref{fig:map_cl}. The architecture consists of three components: (1) an anchor sample representing 3D IMU data captured during in-air drawing of a specific letter, (2) a positive sample containing 2D IMU data of the same letter drawn on a flat surface, and (3) a negative sample comprising 2D IMU data of a different letter drawn on a flat surface.

The pipeline employs two 1D-CNN-based encoders to extract latent embeddings from raw IMU time-series data, with one encoder tailored for 2D data and the other for 3D data.
Following training, the encoder for 3D data is utilized as the similarity estimator in the pipeline, as shown in Figure \ref{fig:overview}.
The network is optimized using a triplet loss function:
$$L = \max(D(g_a, g_p) + margin - D(g_a, g_n), 0),$$
where $D(\cdot)$ denotes the Euclidean distance between embeddings, and $margin$ is a hyper-parameter.

Through this deep metric objective, the framework learns to project 3D data onto 2D data that represents the same letter.
During reference, the similarity estimator first extracts 3D embeddings from the input 3D data. Based on the embeddings, the corresponding projected 2D data is retrieved from the flat-surface gesture gallery and then used for the following letter recognition with LLMs.
When evaluated on the test set, the mapping approach achieved an accuracy of 93.08\%, indicating its efficacy in dimensional reduction by projecting 3D motion patterns to their 2D equivalents.

\begin{table}[t]
\centering
\small
\begin{tabular}{c|cccc}
\hline
\multirow{2}{*}{\textbf{Word Length}} & \multicolumn{4}{c}{\textbf{Number of Samples}} \\
& \textbf{2} & \textbf{3} & \textbf{4} & \textbf{5} \\
\hline
3 & 0.832 & 0.879 & 0.888 & 0.860 \\
4 & 0.795 & 0.780 & 0.790 & 0.795 \\
5 & 0.770 & 0.780 & 0.760 & 0.765 \\
6 & 0.630 & 0.615 & 0.625 & 0.600 \\
\hline
\end{tabular}
\caption{Match accuracy for words formed by multiple letters, across varying word lengths and sample times.}
\label{tab:word_length_samples}
\vspace{-1em}
\end{table}

\subsection{Exp: Contextual Letter Series}\label{sec:contextual_evaluation}
In real-world applications, such as using AR/VR controllers in mid-air, users typically draw letters as components of words rather than as isolated characters.
This experiment evaluates the performance of the fine-tuned LLaMA-3-8B model in interpreting letters within the context of such letter sequences.

\noindent\textbf{Experimental Setup.}
The dataset for this experiment comprised 1,500 common English nouns, specifically selecting words with lengths between 3 and 6 letters for evaluation purposes. For each letter in the selected words, multiple samples were generated by drawing the letter $k$ times, where $k \in [2, 5]$. These samples were subsequently processed by the model again, which aggregated the predictions for individual letters to infer the target word. 

\noindent\textbf{Illustrative Examples.}
We present two examples to demonstrate the experimental process and outcomes. In the first example, the target word ``LAB'' had each letter drawn three times ($k=3$). The individual letter predictions were: $L: \{L, L, V\}$, $A: \{A, N, K\}$, and $B: \{B, B, B\}$. Based on these predictions, the LLM successfully identified the word as ``LAB''. Conversely, the second example illustrates a case where error correction was insufficient. For the ground truth word ``WEST'' with $k=3$, the letter predictions were: first letter: $\{N, W, N\}$; second letter: $\{E, Q, Q\}$; third letter: $\{J, L, S\}$; and fourth letter: $\{T, T, S\}$. In this case, the final prediction was ``NEST'', showing how multiple letter recognition errors can lead to plausible but incorrect word predictions.

\noindent\textbf{Results.}
Table~\ref{tab:word_length_samples} shows the accuracy for varying the word length and number of samples. Interestingly, the results do not demonstrate a consistent correlation between sample quantity and accuracy. However, accuracy noticeably decreases as word length increases from 3 to 6 letters, with accuracy about 78\% for words no longer than 5 characters. These findings suggest two compounding challenges with longer words: first, for letter prediction models (like our fine-tuned LLaMA-3-8B), prediction errors accumulate across multiple characters; second, for word prediction models, the possible word combinations grow exponentially unless the letter prediction achieves near-perfect accuracy. 

%% file: tex_fmsys/07_discussion.tex
\section{Discussion and Future Work}\label{sec: discussion}
\noindent\textbf{Fine-grained human activity recognition.} 
This paper mainly focuses on letter recognition, including both flat-surface and in-air cases, which is a typical kind of fine-grained HAR.
In future work, it could be extended to other tasks beyond letter recognition, such as arm gesture recognition and fine-grained daily activity monitoring, enabling broader application scenarios.

\noindent\textbf{Advantage of context-based prediction.}
In Section~\ref{sec:contextual_evaluation}, we conducted contextual letter series experiments, which show the advantage of LLM-based letter prediction due to its self-correction capability. This advantage could benefit further fine-tuning to progressively improve results, as research by \cite{google2024wearable} has shown that scaling laws also apply to wearable devices, providing more training data yields better model performance. This contextual letter prediction and correction mechanism may facilitate continuous learning for LLMs without requiring labor-intensive human labeling.

\noindent\textbf{Overhead.}
In this work, we treat multi-channel IMU data as pseudo-text input to LLMs, which leads to an unavoidably large number of tokens. We format the data with 3 decimal points to maximize precision while minimizing token count, but this still results in more than 8k total tokens per prompt. For this fine-grained task, we maintained the original sampling rate of the IMU data, as lower rates may cause performance degradation. Future work exploring the relationship between sampling rate and model performance could further enhance the effectiveness of LLMs in IMU-based fine-grained HAR understanding.

%% file: tex_fmsys/06_conclusion.tex
\section{Conclusion}\label{sec: conclusion}
This work explores the use of Large Language Models (LLMs) for fine-grained Human Activity Recognition, specifically focusing on mid-air letter recognition. Our investigation reveals important findings about how well LLMs understand IMU data. Pretrained LLMs initially struggle with direct interpretation of IMU data, but show significant improvements through fine-tuning with a few examples. Small language models achieved a 129$\times$ accuracy increase in 2D letter recognition. To address the challenging 3D recognition task, we developed a mapping approach that converts 3D IMU data into 2D equivalents. Our contextual letter series experiment showed that LLMs can maintain approximately 78\% end-to-end accuracy for words up to 5 characters in length.

%% file: tex_fmsys/ack.tex
\begin{acks}
This research was partially supported by the National Science Foundation under Grant Number CNS-1943396. The views and conclusions contained here are those of the authors and should not be interpreted as necessarily representing the official policies or endorsements, either expressed or implied, of Columbia University, NSF, or the U.S. Government or any of its agencies.
\end{acks}